\documentclass{bioinfo}
\copyrightyear{2016} \pubyear{2016}

\access{Advance Access Publication Date: Day Month Year}
\appnotes{Manuscript Category}

\usepackage{graphicx}
\usepackage{epsf}
\usepackage{bm}
\usepackage{verbatim}
\usepackage{amsmath}
\usepackage{amssymb}
\usepackage{multirow}
\usepackage{url}
\usepackage{caption}
\usepackage{algorithm}
\usepackage{algorithmic}
\usepackage{color}
\graphicspath{{fig/}}
\usepackage{paralist}

\newcommand{\x}{{\mathbf x}}

\newcommand{\cref}[1]{Condition~(\ref{#1})} 

\begin{document}
\firstpage{1}

\subtitle{Subject Section}

\title[DeepChrome]{DeepChrome:  Deep-learning for predicting gene expression from histone modifications.}
\author[R. Singh \textit{et~al}.]{Ritambhara Singh, Jack Lanchantin, Gabriel Robins, and Yanjun Qi\,$^{*}$}
\address{Department of Computer Science, University of Virginia, Charlottesville, VA, U.S.A}

\corresp{$^\ast$To whom correspondence should be addressed.}

\history{}

\editor{This work will be published originally in Bioinformatics Journal at \url{http://bioinformatics.oxfordjournals.org}}

\abstract{\textbf{Motivation:} Histone modifications are among the most important factors that control gene regulation. Computational methods that predict gene expression from histone modification signals are highly desirable for understanding their combinatorial effects in gene regulation. This knowledge can help in developing `epigenetic drugs' for diseases like cancer. Previous studies for quantifying the relationship between histone modifications and gene expression levels either failed to capture combinatorial effects or relied on multiple methods that separate predictions and combinatorial analysis. This paper develops a unified discriminative framework using a deep convolutional neural network to classify gene expression using histone modification data as input. Our system, called DeepChrome, allows automatic extraction of complex interactions among important features. To simultaneously visualize the combinatorial interactions among histone modifications, we propose a novel optimization-based technique that generates feature pattern maps from the learnt deep model. This provides an intuitive description of underlying epigenetic mechanisms that regulate genes. \\
\textbf{Results:} We show that DeepChrome outperforms state-of-the-art models like Support Vector Machines and Random Forests for gene expression classification task on 56 different cell-types from REMC database. The output of our visualization technique not only validates the previous observations but also allows novel insights about combinatorial interactions among histone modification marks, some of which have recently been observed by experimental studies.\\
\textbf{Availability:} Codes and results are available at \url{www.deepchrome.org}\\ 
\textbf{Contact:} \href{yanjun@virginia.edu}{yanjun@virginia.edu}\\
\textbf{Supplementary information:} Supplementary data \footnote{Contains details of data processing, implementation and performance for all the different cell types.} are available at \textit{Bioinformatics} online.}

\maketitle

\section{Introduction}
Gene regulation is the process of controlling gene expression to become high or low. Cells use a wide range of mechanisms to regulate genes and increase or decrease specific gene products through translation such as proteins. Multiple factors combinatorially regulate genes at the DNA level. These can range from mutations in DNA sequences to various proteins binding to them. A principle factor that plays a key role in this transcriptional regulation is the modification of histones. DNA strings are wrapped around ``bead''-like structures called nucleosomes, which are composed of eight histone proteins with DNA wrapped around the proteins. These histone proteins are prone to modifications (e.g. methylation) that can change the spatial arrangement of the DNA. This allows or restricts the binding of different proteins to DNA that leads to different forms of gene regulation. The importance of histone modifications in gene regulation is supported by evidence that aberrant histone modification profiles have been linked to cancer (\cite{bannister2011regulation}). Unlike DNA mutations, the epigenetic changes (like histone modifications) are potentially reversible. This crucial difference makes the study of histone modifications impactful in developing `epigenetic drugs' for cancer treatment.

In this direction, the role of histone modifications in controlling gene expression has been investigated for many years and has resulted in the Histone Code Hypothesis. According to this hypothesis, combinations of different histone modifications specify distinct chromatin (DNA scaffold) states and cause distinct downstream effects, such as gene regulation. Advancement in sequencing technology has allowed us to quantify gene expression and also profile different histone modifications as signals present in regions flanking (i.e surrounding) the gene. Initial studies, like \cite{lim2009defining} and \cite{cain2011gene}, investigated experimentally the correlation between histone modification marks and gene regulation. 

\begin{figure}[t]
\includegraphics[width=\columnwidth]{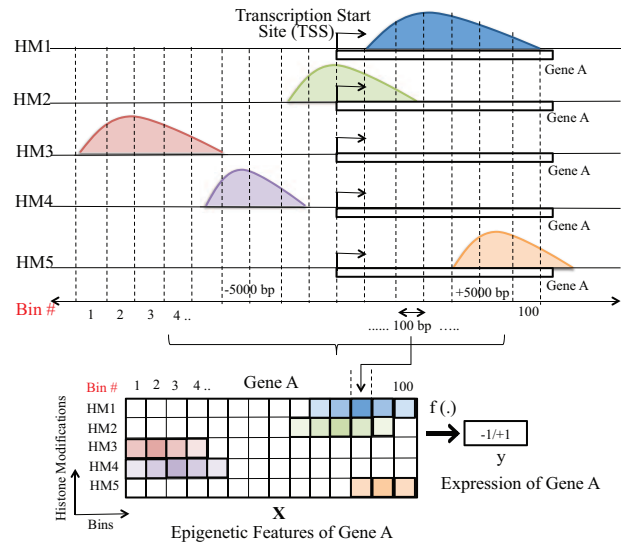}
\caption{\textbf{Feature Generation for DeepChrome model.} Bins of length $100$ base-pairs (bp) are selected from regions ($+/- 5000$ bp) flanking the transcription start site (TSS) of each gene. The signal value of all five selected histone modifications in bins forms input matrix X, while discretized gene expression (label $+1/-1$) is the output y.}\vspace*{-9pt}
\label{fig:featgen}
\end{figure}

Multiple computational models have been proposed to use histone modifications in predicting gene expression (surveyed by \cite{dong2013correlation}). \cite{karlic2010histone} and \cite{costa2011predicting} used linear regression to quantify the relationship between histone modifications and gene expression. This was followed by \cite{cheng2011statistical} using Support Vector Machines (SVMs) for the task of gene expression prediction from histone modification features. Separately, \cite{cheng2011statistical} inferred the pair-wise combinatorial contribution of different histone modifications as binary interaction terms among features. Furthermore, they studied higher order relationships using Bayesian networks. Next, \cite{dong2012modeling} introduced Random Forests for predicting gene expression from histone modification marks. The authors studied the combinatorial effects by dividing histone modification marks into four functional categories and then reported the influence of these categorical combinations through prediction performance. Recently, \cite{ho2015combinatorial} introduced a rule-based learning model and reported 83 rules that capture the interaction effects of different histone modification marks on gene regulation. 
There are a few drawbacks in the previous studies. First, they rely on multiple models to separate prediction and combinatorial analysis. Second, for input features, some of them take the average value of histone modification signal from the gene region (\cite{karlic2010histone,costa2011predicting}) and fail to capture the subtle differences among signal distributions of histone modifications. To overcome this issue, most of the later methods use a `binning' approach, that is, a large region surrounding the gene transcription start site (TSS) is converted into consecutive smaller bins. These studies either have separate models for each bin (bin-specific strategy in \cite{cheng2011statistical}) or select the most relevant bins (best-bin strategy in \cite{dong2012modeling}) as the model input, and therefore cannot model connections among input bins. Furthermore, when performing combinatorial analysis among histones, most previous studies use the best-bin strategy and fail to model the representation of neighboring bins. As seen in Figure~\ref{fig:featgen}, histone modification signals can span across multiple neighboring local bins. 

Recently, deep learning methods have achieved state-of-the-art accuracy on many prediction tasks such as image classification (\cite{krizhevsky2012imagenet}). A deep learning model automatically learns complex functions that map inputs to outputs. It eliminates the need to use hand-crafted features or rules. One such variant of deep learning is called Convolutional Neural Networks (CNNs), which capture both local and global representations in the input samples to learn the most important features that, in turn, help make better predictions. CNNs have been used successfully in computer vision (\cite{pinheiro2013recurrent,szegedy2015going}), natural language processing (\cite{kim2014convolutional,collobert2008unified}) and bioinformatics (\cite{alipanahi2015predicting,zhou2015predicting}).

This paper introduces DeepChrome, a unified CNN framework that automatically learns combinatorial interactions among histone modification marks to predict the gene expression. It is able to handle all the bins together, capturing both neighboring range and long range interactions among input features, as well as automatically extract important features. In order to interpret what is learned, and understand the interactions among histone marks for prediction, we also implement an optimization-based technique for visualizing combinatorial relationships from the learnt deep models. Through the CNN model, DeepChrome incorporates representations of both local neighboring bins as well as the whole gene TSS flanking regions, therefore overcoming the challenges faced by previous studies.
 
The contributions of this work can be summarized as follows: 
\begin{itemize}
\item DeepChrome is the first deep learning implementation for gene expression prediction task using histone modification data as feature inputs. We apply our model on histone modification signal data for 56 different cell types from latest REMC database (\cite{kundaje2015integrative}).
\item Our model outperforms previous state-of-the-art SVM and Random Forest implementations for 56 prediction tasks.
\item DeepChrome enables visualization of high-order combinatorial relationships among different histone modification signals. The findings from our experiments not only validate previous observations but also provide insights supported by recent biological evidence in literature. \vspace*{-9pt}
\end{itemize}

\begin{table*}[!t]
\centering
\processtable{Comparison of previous studies for the task of quantifying gene expression using histone modification data. The columns indicate properties (a) whether the study has a unified end-to-end architecture or not (b) if it captures non-linearity among features (c) how has the bin information been incorporated (c) if representation of features is modeled on local and global scales (d) whether gene expression prediction is provided and finally, (e) if combinatorial interactions among histone modifications are modeled. DeepChrome is the only model that exhibits all six desirable properties.\label{Tab:03}}{\begin{tabular}{@{}lcp{15mm}p{15mm}p{20mm}p{15mm}p{15mm}p{15mm}p{15mm}@{}}\toprule  & Computational Study &
Unified Strategy & Non-linear model & Including Bin Info & \multicolumn{2}{c}{Representation Learning} & Prediction & Combinatorial Interactions \\\midrule
 &  &  & & & Neighboring bins & Whole Region &  & \\\midrule
 & Linear Regression (\cite{karlic2010histone}) & $\times$ & $\times$ & $\times$ & $\times$ & $\checkmark$ & $\checkmark$ & $\times$  \\
 & Support Vector Machine (\cite{cheng2011statistical}) & $\times$ & $\checkmark$ & Bin-specific strategy & $\times$ & $\checkmark$ & $\checkmark$ & $\checkmark$ \\
 & Random Forest (\cite{dong2012modeling}) & $\times$ & $\checkmark$ & Best-bin strategy & $\times$ & $\checkmark$  & $\checkmark$ & $\times$ \\
 & Rule Learning (\cite{ho2015combinatorial}) & $\times$ & $\checkmark$ & $\times$ & $\times$ & $\checkmark$ & $\times$ & $\checkmark$ \\
 & \textbf{DeepChrome} & $\checkmark$ & $\checkmark$ & Automatic & $\checkmark$ & $\checkmark$ & $\checkmark$ & $\checkmark$ \\\botrule 
\end{tabular}}{}\vspace*{-9pt}
\end{table*}

\section{Related Work}

The combinatorial effect of histone modifications in regulating gene expression has been studied throughout literature (\cite{dong2013correlation}). To better understand this relationship, scientists have generated experimental datasets quantifying gene expression and histone modification signals across different cell-types. These datasets have been made available through large-scale repositories, the latest being the Roadmap Epigenome Project (REMC) (\cite{kundaje2015integrative}). \vspace*{-12pt}

\subsection{Computational methods for predicting gene expression using histone modifications}

Computational methods have shown initial success in modeling and understanding interactions among chromatin features, such as histone modification marks, to predict gene expression.
\cite{karlic2010histone} established that there exists a quantitative relationship between histone modifications and gene expression. They applied a linear regression model on histone modification signals and predicted gene expression from human T-cell studies (\cite{wang2008combinatorial}). They reported a high correlation of their predictions with the observed gene expressions (Pearson coefficient $r=0.77$) and showed that a combination of only two to three specific modifications is sufficient for making accurate predictions. Extending this concept further, \cite{costa2011predicting} implemented a mixture of several linear regression models to extract the relative importance of each histone modification signal and its effect on gene expression (high or low). This study confirmed the activator and repressor roles of H3K4me3 and H3K27me3 respectively. It also demonstrated that a mixture of two regression models performs better than a single regression model. Both these studies applied relatively simple modeling on a small dataset. They used the mean signal of the whole transcription start site (TSS) flanking regions as input features. This leads to a potential bias since histone modification signals exhibit diverse patterns of local distributions with regard to different genes. Ignoring the details of these neighborhood patterns is undesirable.

\cite{cheng2011statistical} applied Support Vector Machine (SVM) models on worm datasets (\cite{celniker2009unlocking}) and reformulated the task as gene expression classification and prediction. The authors divided regions flanking transcription start site (TSS) and transcription termination site (TTS) into 100 base-pair (bp) bins and used the histone modification signal in each bin as a feature for the SVM. To incorporate information from all positions or bins, they trained different models for different bins that resulted in 160 models for 160 bins. They validated the existence of the quantitive relationship between histone modifications and gene expression by such bin-specific modeling. Furthermore, using a separate linear regression model, the paper inferred pair-wise interactions between different histone modifications using binary combinatorial terms. Since it is infeasible to consider all possible higher order interaction terms through polynomial regression, Bayesian networks were then used for modeling such relationships. However, Bayesian networks do not take into consideration local neighboring bin information and their highly connected output network is difficult to interpret.

Using a similar experimental setup, \cite{dong2012modeling}, applied a Random Forest Classifier on histone modification signals to classify gene expression as high or low. They then used the classified outputs as inputs to a linear regression model to predict the gene expression value. They used human datasets across 7 different cell types (\cite{encode2012integrated}) and reported a high correlation (Pearson coefficient $r=0.83$) between predicted and actual gene expressions. To include information from all bins into a single model, the authors performed feature selection where only the bin value which correlated the most with gene expression was used as input. For combinatorial analysis, instead of studying all possible combinations, the 11 histone modifications were grouped into four functional categories. These groupings were used to determine prediction accuracy based on each category as a sole feature as well as combinations of different categories. This technique gives a broader picture of the combinatorial effect. However, individual details of histone modifications are missed. In addition, this approach cannot capture the possible influence of other bins besides the ``best bin'' for gene regulation.

In order to to elucidate the possible combinatorial roles of histone modifications in gene regulation, \cite{ho2015combinatorial} applied rule learning on the T-cells datasets (\cite{wang2008combinatorial}) and produced 83 valid rules for gene expression (high) and repression (low). The authors selected the 20 most discriminative histone modifications as input into a rule learning system. They used several heuristics to filter out unexpected rules that were obtained by the learning system after scanning the entire search space. However, this study does not consider detailed feature patterns across local bins and does not perform prediction of gene expression.

\cite{ernst2015large} leveraged the correlated nature of epigenetic signals in the REMC database, including histone modifications. Their tool, ChromImpute, imputed signals for a particular new sample using an ensemble of regression trees on all the other signals and samples. EFilter (\cite{kumar2013uniform}), a linear estimation algorithm, predicted gene expression in a new sample by using imputed expression levels from similar samples. Unlike the studies discussed above, these works focus on imputing or predicting signals for new samples. 

In summary, we compare the aforementioned studies in Table~\ref{Tab:03} using six different functional aspects. All previous studies have missed one or more aspects. In contrast, our model, DeepChrome, exhibits all six properties. It is a unified framework, scalable to large datasets. It performs automatic feature selection and can incorporate information from all the bin positions. It also provides an optimization-based strategy to simultaneously visualize combinatorial relationships among multiple histone modifications. \vspace*{-12pt}

\subsection{Connecting to deep learning}

\begin{figure*}[t]
\centering
\includegraphics[width=\linewidth]{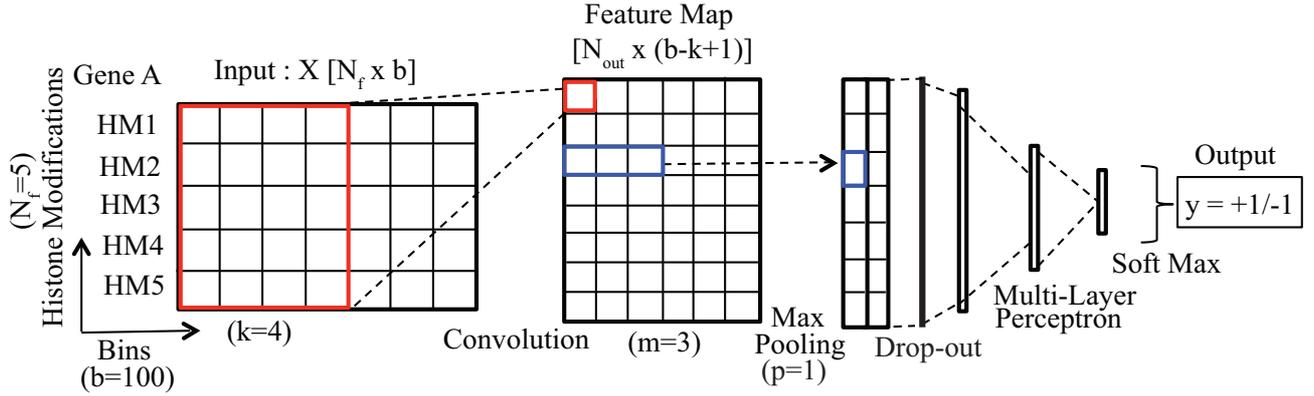}
\caption{\textbf{DeepChrome convolution neural network (CNN) model}. The input matrix X, comprising of 100 bins with signals from five histone modifications, goes through different CNN stages. These stages are : convolution, pooling followed by dropout, and multi-layer perceptron with alternating linear and non-linear layers. Softmax function, in the end, maps the output from the model into classification prediction.} \vspace*{-9pt}
\label{fig:model}
\end{figure*} 

In recent years, deep learning models have become popular in the bioinformatics community, owing to their ability to extract meaningful and hierarchical representations from large datasets. \cite{qi2012unified} used a deep multi-layer perceptron (MLP) architecture with multitask learning to perform sequence-based protein structure prediction. \cite{zhou2014deep} created a generative stochastic network to predict secondary structure on the same data as used by \cite{qi2012unified}. Recently, \cite{lin2016must} outperformed all the state-of-the-art works for protein property prediction task by using a deep convolutional neural network architecture. \cite{leung2014deep} implemented a deep neural network for predicting alternative splicing patterns in individual tissues and differences of splicing patterns across tissues. Later, \cite{alipanahi2015predicting} applied a convolutional neural network model for predicting sequence specificities of DNA-and RNA-binding proteins as well as generating motifs, or consensus patterns, from the features that were learnt by their model. \cite{lanchantin2016demo} proposed a deep convolutional/highway MLP framework for the same task and demonstrated improved performance. Similarly, \cite{zhou2015predicting} used DNA sequences as inputs to predict different chromatin features and understand the effect of non-coding variants on these measurements of interest. In contrast, to our knowledge, a deep learning framework has not yet been explored for the task of understanding the combinatorial effect of histone modifications on gene regulation. \vspace*{-9pt}

\section{Approach}
Previous computational methods failed to capture higher-order combinatorial effects among histone modifications, used bin related strategies that cannot represent neighboring bins, or relied on multiple methods to separate prediction and combinatorial analysis. We utilize a deep convolutional neural network model for predicting gene expression from histone modification data. The network automatically learns both the combinatorial interactions and the classifier jointly in one unified discriminative framework, eliminating the need for human effort in feature engineering. Since the combinatorial effects are automatically learned through multiple layers of features, we present a visualization technique to extract those interactions and make the model interpretable. \vspace*{-9pt}

\subsection{Input Generation}
Aiming to systematically understand the relationship between gene regulation and histone modifications, we divided the $10,000$ basepair (bp) DNA region ($+/- 5000$ bp) around the transcription start site (TSS) of each gene into bins of length $100$ bp. Each bin includes $100$ bp long adjacent positions flanking the TSS of a gene. In total, we consider five core histone modification marks from REMC database (\cite{kundaje2015integrative}), which are summarized in Table~\ref{Tab:01}. These five histone modifications are selected as they are uniformly profiled across all cell-types considered in this study. This makes the input for each gene a $5\times100$ matrix, where columns represent different bins and rows represent histone modifications. For each bin, we report the value of all 5 histone signals as the input features for that bin (Figure~\ref{fig:featgen}). We formulate the gene expression prediction as a binary classification task. Specifically, the outputs of DeepChrome are labels $+1$ and $-1$, representing gene expression level as high or low, respectively. Following \cite{cheng2011statistical}, we use the median gene expression across all genes for a particular cell-type as a threshold to discretize the gene expression target. Figure~\ref{fig:featgen} summarizes our input matrix generation strategy.

Our setup is similar to \cite{cheng2011statistical} and \cite{dong2012modeling}, except that we primarily focus on the regions around TSS instead of also including regions from gene body or transcription termination site (TTS). This is based on the observations from \cite{cheng2011statistical} showing that signals close to the TSS are the most informative, therefore eliminating the need to obtain bins from regions toward the end of the gene. In addition, due to the scalability of CNNs, we were able to use larger regions flanking TSS than previous studies in order to better capture effects of distal signals as well as to cover more regions. This therefore enhances the possibility to model long range interactions among histone modifications. \vspace*{-9pt}

\begin{table}[!t]
\processtable{Five core histone modification marks, as defined by \cite{kundaje2015integrative}, along with their functional categories\label{Tab:01}} {\begin{tabular}{@{}llll@{}}\toprule Histone Mark &
Associated with & Functional Category\\\midrule
H3K4me3 &Promoter regions & Promoter mark\\
H3K4me1 & Enhancer regions & Distal mark\\
H3K36me3 & Transcribed regions & Structural mark\\
H3K9me3 & Heterochromatin regions & Repressor mark\\
H3K27me3 & Polycomb repression & Repressor mark\\\botrule
\end{tabular}}{} \vspace*{-9pt}
\end{table}

\subsection{An end-to-end architecture based on Convolutional Neural Network (CNN)}
Convolution Neural Networks (CNNs) were first popularized by \cite{lecun1998gradient} and have since been extensively used for a wide variety of applications. 
In this paper, we have implemented a CNN for gene expression classification task using the Torch7 (\cite{collobert2011torch7}) framework. Our DeepChrome model, summarized in Figure~\ref{fig:model}, is composed of five stages. We assume our training set contains $N_{samp}$ gene samples of the labeled-pair form $(\mathbf{X}^{(n)},y^{(n)})$, where $\mathbf{X}^{(n)}$ are matrices of size $N_{f} ($=5$) \times b$ ($=100$) and $y^{(n)} \in \{-1,+1\}$ for $n \in \{1,...,N_{samp}\}$.
\begin{enumerate}
\item \textbf{Convolution:} We use temporal convolution with $N_{out}$ filters, each of length $k$. This performs a sliding window operation across all bin positions, which produces an output feature map of size $N_{out} \times (b-k+1)$. Each sliding window operation applies $N_{out}$ different linear filters on $k$ consecutive input bins from position $p = 1$ to $(b-k+1)$. In Figure~\ref{fig:model}, the red rectangle shows a sliding window operation with $k=4$ and $p=1$. Given an input sample $\mathbf{X}$ of size $N_{f}\times{b}$, the feature map, $\mathbf{Z}$, from convolution is computed as follows : \vspace*{-2pt}

\begin{equation}
\begin{aligned}
\mathbf{Z} &= f_{conv}(\mathbf{X})\\
\mathbf{Z}_{p,i} & = \mathbf{B}_i + \sum_{j=1}^{N_{f}} \sum_{r=1}^{k} \mathbf{W}_{i,j,r} \mathbf{X}_{p+r-1,j}
\end{aligned}
\end{equation}
\vspace*{-2pt}
This is generated for the $p^{th}$ sliding neighborhood window and the $i^{th}$ hidden filter, where $p \in \{1,...,(b-k+1)\}$ and $i \in \{1,...,N_{out}\}$.
$\mathbf{W}$, of size $N_{out} \times N_{f} \times k$, and $\mathbf{B}$, of size $N_{out} \times 1$, are the trainable parameters of the convolution layer and $N_{out}$ denotes the number of filters. 

\item \textbf{Rectification:} In this stage, we apply a non-linearity function called rectified linear unit (ReLU). The ReLU is an element-wise operation that clamps all negative values to zero: \vspace*{-2pt}
\begin{equation}
f_{relu}(z)=\mathrm{relu}(z) = \max(0, z)
\end{equation} \vspace{-2pt}

\item \textbf{Pooling:} Next, in order to learn translational invariant features, we use temporal maxpooling on the output from the first two steps. Maxpooling simply selects the max values in a certain range, which forms a smaller representation of a large TSS-proximal region for a given gene. Maxpooling is applied on an input $\mathbf{Z}$ of size $N_{out}\times P$, where $P=(b-k+1)$. With a pooling size of $m$, we obtain an output $\mathbf{V}$ of size $N_{out}\times \lfloor\frac{P}{m}\rfloor$:\vspace*{-2pt}
\begin{equation}
\begin{aligned}
\mathbf{V} &= f_{maxpool}(\mathbf{Z})\\
\mathbf{V}_{i,p} &= \max_{j=1}^m { \mathbf{Z}_{i,m(p-1)+j} }
\end{aligned}
\end{equation}
\vspace{-2pt}
where $p \in \{1,...,\lfloor\frac{P}{m}\rfloor\}$ and $i \in \{1,...,N_{out}\}$. In Figure~\ref{fig:model}, the blue rectangle shows the result of a maxpooling operation on the feature map where $m=3$.
\item \textbf{Dropout:} The output is then passed though a dropout layer (\cite{srivastava2014dropout}), which randomly zeroes the inputs to the next layer during training with a chosen probability of $0.5$. This regularizes the network and prevents over-fitting. It resembles ensemble techniques, like bagging or model averaging, which are very popular in bioinformatics.
\item \textbf{Classical feed-forward neural network layers:} Next, the learnt region representation is fed into a multi-layer perceptron (MLP) classifier to learn a classification function mapping to gene expression labels. This standard and fully connected multi-layer perceptron network has multiple alternating linear and non-linear layers. Each layer learns to map its input to a hidden feature space, and the last output layer learns the mapping from a hidden space to the output class label space ($+1/-1$) through a softmax function. 

Figure~\ref{fig:model}, shows a MLP with 2 hidden layers and a softmax function at the end. This stage is represented as $f_{mlp}(.)$. \vspace*{-5pt}
\end{enumerate}
The whole network output form can be written as:
\begin{equation}
f(\mathbf{X}^{(n)})=f_{mlp}(f_{maxpool}(f_{relu}(f_{conv}(\mathbf{X}^{(n)})))) \label{eq:cnn}
\end{equation}
All the above stages are effective techniques that are widely practiced in the field of deep learning. All parameters, denoted as $\Theta$, are learned during training in order to minimize a loss function which captures the difference between true labels $y$ and predicted scores from $f(.)$.\footnote{When training this deep model, parameters, at first, are randomly initialized and input samples are fed through the network. The output of this network is a score prediction associated with a sample. The difference between a prediction output $f(\mathbf{X})$ and its true label $y$ is fed back into the network through a `back-propagation' step.} The loss function $L$, on the entire training set of size $n$, is defined:
\begin{equation}
L=\sum_{n=1}^{N_{samp}} loss(f(\mathbf{X}^{(n)}),y^{(n)}) \label{eq:loss}
\end{equation}

We use stochastic gradient descent (SGD) (\cite{bottou2004stochastic}) to train our model via backpropagation. For a set of training samples, instead of calculating the true gradient of the objective using all training samples, SGD calculates the gradient per sample and updates accordingly on each training sample. For our objective function, the loss $L(.)$ (equation~\ref{eq:loss}) is minimized by the gradient descent step that is applied to update network parameters $\Theta$ as follows:
\begin{equation}
\Theta \leftarrow \Theta - \eta \frac{\partial L}{ \partial \Theta}   
\end{equation} 
where $\eta$ is the learning rate (set to 0.001).

\subsection{Visualizing combinatorial effect through optimization}

In addition to being able to make high accuracy predictions on the gene expression task, an important contribution of DeepChrome is that it allows us to discover and visualize the combinatorial relationships between different histone modifications which lead to such predictions. Until recently, deep neural networks were viewed as ``black boxes'' due to the automatically learned features spanning multiple layers. Since gene expression is dependent on the combinatorial interactions among histone modifications, it is critical to understand how the network extracts features and makes its predictions. In other words, we wish to understand the combinatorial patterns of histone modifications which lead to either a high or low gene expression prediction by the network. We attempt to do this by extracting a map of feature patterns that are highly influential in predicting gene expression directly from the trained network. This approach, called a network-centric approach (\cite{yosinski2015understanding}), finds the \textit{class specific} features from the trained model and is independent of specific testing samples. 

The technique we use to generate this visualization was inspired from works by \cite{simonyan2013deep} and 
\cite{yosinski2015understanding}, which seek to understand how a convolutional neural network interprets a certain image class on the task of object detection. We, instead, seek to find how our network interprets a gene expression class (high or low).  Given a trained CNN model and a label of interest ($+1$ or $-1$) in our case, we perform a numerical optimization procedure on the model to generate a feature pattern map which best represents the given class. This optimization includes four major steps:

\begin{enumerate}
\item Randomly initialize an input $\mathbf{X}_{c}$ (of size $N_{f} (=5) \times{b} (=100)$).
\item Find the best values of entries in $\mathbf{X}_{c}$ by optimizing the following equation(\ref{eq:opt}). We search for  $\mathbf{X}_{c}$ so that the loss function is minimized with respect to the desired labels $+1$ (gene expression = high) or $-1$ (gene expression = low). Using equation~(\ref{eq:cnn}), $f(\mathbf{X}_{c})$ provides the predicted label using the trained DeepChrome model on an input $\mathbf{X}_{c}$. We would like to find an optimal feature pattern, $\mathbf{X}_{c}$, so that its predicted label $f(\mathbf{X}_{c})$ is close to the desired class label $c$:
\begin{equation}
\operatorname*{arg\,min}_{X_{c}} L_{visual} = \operatorname*{arg\,min}_{X_{c}} \{L(f(\mathbf{X}_{c}),y=c) + \lambda \lVert \mathbf{X}_{c} \rVert_{2}^{2}\}\label{eq:opt}
\end{equation}
where $c=+1$ or $-1$, $L(.)$ is the loss function defined in equation~(\ref{eq:loss}). $L_{2}$ regularization, $\lVert \mathbf{X}_{c} \rVert_{2}^{2}$, is applied to scale the signal values in $\mathbf{X}_{c}$, and $\lambda$ is the regularization parameter.
A locally-optimal $X_{c}$ can be found by the back-propagation method. This step is similar to the CNN training procedure, where back-propagation is used to minimize the loss function by optimizing the \textit{network} parameters $\Theta$. However, in this case, the optimization is performed with respect to the \textit{input values} ($\mathbf{X}_{c}$) and the network parameters are fixed to the values obtained from the classification training. $\mathbf{X}_{c}$ is optimized in the following manner:
\begin{equation}
\mathbf{X}_{c}^{t+1} \leftarrow \mathbf{X}_{c}^{t}- \alpha \frac{\partial L_{visual}}{ \partial \mathbf{X}_{c}}
\end{equation} 
where $\alpha$ is the learning rate parameter and $t$ represents the iteration step of the optimization.
\item Next, we set all the negative output values to $0$ and normalize $\mathbf{X}_{c} \in [0,1]$:
\begin{equation}
\mathbf{X}_{c(norm)} = \frac{\mathbf{X}_{c}}{max(\mathbf{X}_{c})}
\end{equation}
\item Finally, we set a threshold of $0.25$ to define ``active'' bins. Bins in $\mathbf{X}_{c(norm)}$ with values $>0.25$ are considered important since they indicate that such histone modification signals are important for predicting particular class. We count the frequency of these active bins for a particular histone modification mark. A high frequency count ($>$ mean frequency count across all histone modification marks) of active bins indicates the important influence of these histone modification signals on target gene expression level. \vspace*{-1pt}
\end{enumerate}

This visualization technique represents the notion of a class that is learnt by the DeepChrome model and is not specific to a particular gene. The optimized feature pattern map $\mathbf{X}_{c(norm)}$ is representative for a particular gene expression label of $+1$ (high) or $-1$ (low). In Figure~\ref{fig:dashboard}, DeepChrome visualizes $\mathbf{X}_{c(norm)}$ as heat-maps. Through these maps, we obtain intuitive outputs for understanding the combinatorial effects of histone modifications on gene regulation. \vspace*{-9pt}

\section{Experiment Setup}

\begin{table}[!t]
\processtable{Results on \textbf{validation set} (6601 genes) during tuning across different combinations of kernel size $k$ and pool size $m$. $k$ captures the local neighborhood representations of bins and $m$ combines the important representations across whole regions for our CNN model. We report the maximum, minimum and mean AUC score obtained across 56 cell types (or tasks). The best performing values of $k=10$ and $m=5$ (highest Max. and Min. AUC scores) were selected evaluating test performance of DeepChrome. \label{Tab:02}} {\begin{tabular}{@{}lcccc@{}}\toprule  & Kernel Size, Pool Size ($k, m$) & \multicolumn{3}{c}{AUC Scores (Validation Set)} \\\midrule
 & & Max & Min & Mean \\\midrule
 & (5,2) & 0.94 & 0.65 & 0.77 \\
 & (5,5) & 0.94 & 0.65 & 0.77 \\
 & (10,2) & 0.94 & 0.65 & 0.76 \\
 & (10,5) & 0.94 & 0.66 & 0.77 \\\botrule
\end{tabular}}{}\vspace*{-9pt}
\end{table}

\subsection{Dataset}
We downloaded gene expression levels and signal data for five core histone modification signals for 56 different cell types from the REMC database (\cite{kundaje2015integrative}). REMC is a public resource of human epigenomic data produced from hundreds of cell-types. Core histone modification marks, as defined by \cite{kundaje2015integrative}, have been listed in Table~\ref{Tab:01} and are known to play important roles in gene regulation. We focus on these ``core'' histone modifications as they have been uniformly profiled for all 56 cell types through sequencing technologies. The gene expression data has been quantified for all annotated genes in the human genome and has been normalized for all 56 cell types in the REMC database. As mentioned before, the target problem has been formulated as a binary classification task. Thus, each gene sample is associated with a label $+1/-1$ indicating whether gene expression is high or low respectively. The gene expression values were discretized using the median of gene expressions across all genes for a particular cell-type. \vspace*{-9pt}

\begin{figure}[t]
\centering
\includegraphics[width=\columnwidth]{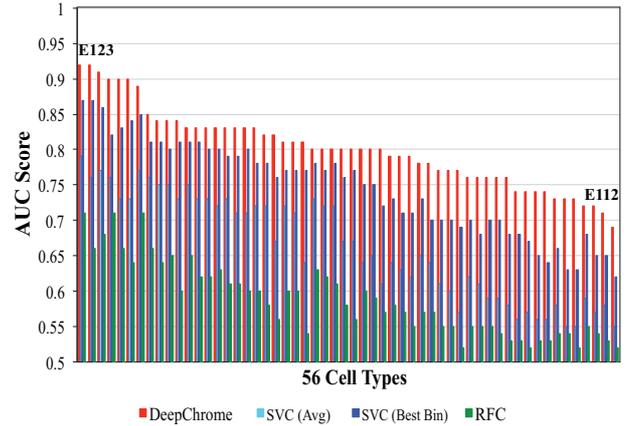}
\caption{\textbf{Performance Evaluation on Test Set.} (Best viewed in color) The bar graph represents AUC scores of DeepChrome versus state-of-the-art baseline models for 56 cell types (i.e. 56 different classification tasks). The results have been arranged from best performing cell type (E123) to the worst performing cell type (E112) for the test set (6600 genes). DeepChrome (Average AUC = 0.80) consistently outperforms both SVM (Average AUC: SVM Best Bin = 0.75 and SVM Avg = 0.66) and Random Forest Classifier (Average AUC = 0.59 ) for the task of binary classification of gene expression. SVM based baseline has a separate model for each bin (bin specific model), thus, results for both average AUC scores across all bins (SVM Avg) and best performing AUC score among the bins (SVM Best Bin) are presented.}\vspace*{-9pt}
\label{fig:perf}
\end{figure}

\subsection{Baselines}
We compare DeepChrome to two baseline studies, \cite{cheng2011statistical} which uses Support Vector Machines (SVM) and \cite{dong2012modeling} which uses a Random Forest Classifier. Their implementation strategies are as follows: 
\begin{itemize}
\item \textit{SVM (\cite{cheng2011statistical}):} The authors selected 160 bins from regions flanking the gene TSS and TTS. Each bin position uses a separate SVM classification model, resulting in 160 different models in total. This gave insights into important bin positions for classifying gene expression as high or low. Following this bin-specific model strategy, we provide results for performance of the best bin (SVM Best Bin) along with average performance across all bins (SVM Avg) in Section 5.1 and in Figure~\ref{fig:perf}.
\item \textit{Random Forest Classifier (\cite{dong2012modeling}):} In this study, bins were selected from regions flanking the TSS, TTS, and gene body. This study selected the bin values having the highest correlation with gene expression as ``best bins''. A matrix with all genes and best bins for each histone modification signal was used as input into the model to predict gene labels ($+1/-1$) as output. Since this baseline performs feature selection using the best bin strategy, our experiment uses the best-bin Random Forest performance as a baseline in Figure~\ref{fig:perf}. \vspace*{-1pt}
\end{itemize}
We implemented these baselines using the python based scikit-learn (\cite{scikit-learn}) package. \vspace*{-9pt}

\begin{figure*}[t]
\centering
\includegraphics[width=\linewidth]{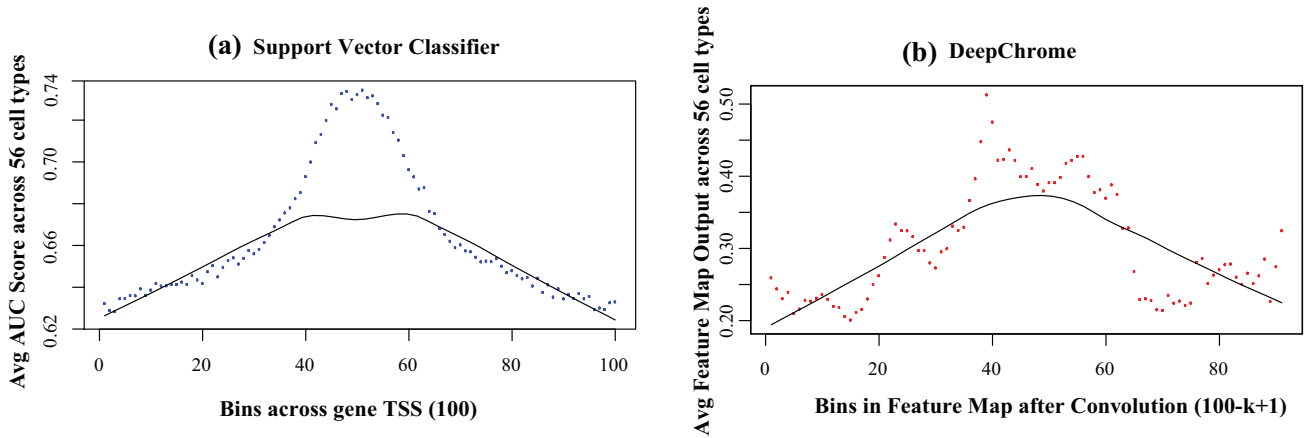}
\caption{\textbf{Validating the influence of positions for gene expression classification.} \cite{cheng2011statistical} reported that the bin positions closer to the transcription start site (TSS) of each gene are more important when predicting gene expression. This is confirmed by our implementation of this bin-specific baseline model in (a). For each bin position, it shows the mean AUC score across all the cell types. In (b) we plot the filter outputs from the convolution layer of DeepChrome model. For each bin, its value has been averaged across all filters and cell-types. The solid lines represent the best-curve fit to the data points plotted in the figures. The trends for both (a) and (b) are similar.} \vspace*{-9pt}
\label{fig:bins}
\end{figure*}

\subsection{Hyperparameter tuning}
For each cell type, our sample set of total 19802 genes was divided into 3 separate, but equal size folds: training (6601 genes), validation (6601 genes) and test (6600 genes) sets. We trained DeepChrome using the following hyperparameters: filter size ($k=\{10, 5\}$), number of convolution filters ($N_{out}=\{20, 50, 100\}$) and pool size for maxpooling ($m=\{2, 5\}$). Table~\ref{Tab:02} presents the validation set results for tuning different combinations of kernel size $k$ and pool size $m$. $k$ denotes the local neighborhood representations of flanking bins. $m$ represents selected whole regions in our CNN model. We report the maximum, minimum and mean AUC scores obtained across 56 cell types. Performances of models using these different hyperparameter values did not vary significantly ($p$-value$\sim0.92$) from each other. We also trained a deeper model with 2 convolution layers and observed no significant ($p$-value$=0.939$) increase in performance. 
\begin{itemize}
\item We selected  $k=10$, $N_{out}=50$, and $m=5$ for training the final CNN models based on highest Max. and Min. AUC scores in Table~\ref{Tab:02}. The number of hidden units chosen for the two multilayer perceptron layers were $625$ and $125$, respectively. We trained the model for 100 epochs and observed that it converged early (around 15-20 epochs). 
\item For the SVM implementation, an RBF kernel was selected and the model was trained on varying hyperparameter values of $C \in \{0.01,0.1,1,10,100,1000\}$ and $\gamma \in \{0.01,0.1,1,2\}$. The $C$ parameter balances the trade-off between misclassification of training examples and simplicity of the decision surface, while the $\gamma$ parameter defines the extent of influence of a single training sample. 
\item For the Random Forest Classifier implementation, we varied the number of decision trees, $n_d \in \{10,20,...,200\}$ trained in each model. 
\end{itemize} \vspace*{-1pt}
All the above models were trained on the training set, and the parameters for testing were selected based on their results on the validation set. We then applied the selected models on the test dataset. The AUC scores \footnote{Area Under Curve (AUC) score from Receiver Operating Characteristic (ROC) curve is interpreted as the probability that a randomly selected ``event'' will be regarded with greater suspicion (in terms of its continuous measurement) than a randomly selected ``non-event''.  AUC score ranges between 0 and 1, where values closer to 1 indicate more successful predictions.} (performance metric) are reported in Section 5.1. \vspace*{-9pt}

\section{Results}

\begin{figure*}[t]
\centering
\includegraphics[width=\linewidth]{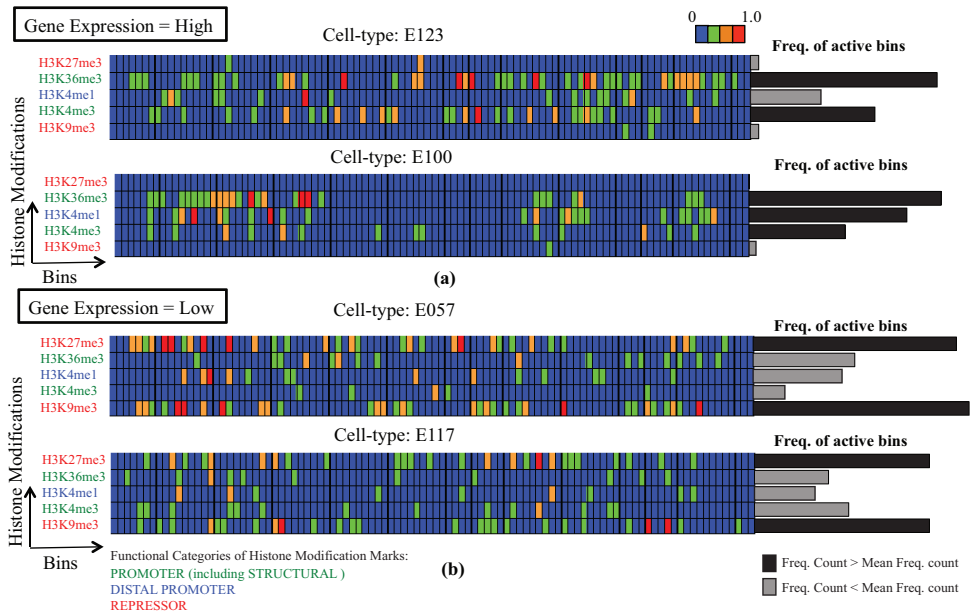}
\caption{\textbf{DeepChrome visualization.} (Best viewed in color) Four examples of feature maps generated by our optimization technique from four trained models. The scores in these feature maps are $ \in [0,1]$ and a threshold of $0.25$ was selected to indicate ``active'' (or important) bins. The bar graph represents the count of active bins for each histone modification. Higher frequency count ($>$ mean frequency count across all histone marks) indicates greater influence of the histone modification mark in prediction of gene expression labels. Multiple marks with high frequency count are considered to be combinatorially affecting the gene expression to become high or low. (a) As expected, we observe a relationship among promoter and structural histone modification marks (H3K4me3 and H3K36me3) when gene expression is high. (b) Similarly, we observe an opposite trend with repressor marks (H3K9me3 and H3K27me3) showing combinatorial relationship, when gene expression is low. These pattern maps not only support previous quantitative observations in \cite{cheng2011statistical} and \cite{dong2012modeling}, but also provide novel insights that are supported by recent biological studies. For example, a recent study by \cite{boros2014polycomb} has reported evidence of coexistence of H3K27me3 and H3K9me3 modifications in gene silencing.} \vspace*{-9pt}
\label{fig:dashboard}
\end{figure*}

\subsection{Performance Evaluation}
The bar graph in Figure~\ref{fig:perf} compares the performance of DeepChrome and three baselines on test data set for gene expression classification across 56 different cell-types (or tasks). DeepChrome (Average AUC = 0.80) outperforms the baselines for all the cell types shown along the X-axis. As discussed earlier, \cite{cheng2011statistical} implement a different SVM model for each bin position. Therefore, we report both average AUC score for all the bins (SVM Avg) as well as the best AUC score among all bins (SVM Best Bin). `SVM Best Bin' (Average AUC = 0.75) gives better results than `SVM Avg' (Average AUC = 0.66). However, its AUC scores are still lower than those of DeepChrome. Random Forest gives the worst performance (Average AUC =0.59). Additionally, we observe that the performances of all three models vary across different cell types and follow a similar trend. For some cell types, like E123, the prediction task resulted in higher AUC scores among all models than other cell types. \vspace*{-9pt}
	
\subsection{Validating the influence of bin positions on prediction}
\cite{cheng2011statistical} obtained predictions for each bin (due to bin-specific strategy) and reported that, on average, the best AUC scores were obtained from bins that are close to the TSS. Figure~\ref{fig:bins} (a) shows that our implementation of this SVM baseline confirms this observation. Since our convolutional network makes a prediction on the entire flanking region (i.e. all the bins at once), we cannot evaluate the AUC for each individual bin. However, we can roughly determine which bins are the most influential for a specific gene prediction. To do this, we look at the strongest activations among the output of the convolution step (the feature map, as shown in Figure~\ref{fig:model}). Since the column in the feature map corresponds to the bins in the input region, we can simply look at the feature map values to determine which bin positions are most influential for that prediction. To validate our model, we ran all of our test samples through a trained deep network, and took the average of all the feature maps across all 56 models. Figure~\ref{fig:bins}(b) shows that bins near the center, closer to TSS, are assigned with higher values by the convolution layers. This indicates that DeepChrome maintains similar trends as observed by \cite{cheng2011statistical}. This trend indicates histone modification signals of bins that are closer to TSS are more influential in gene predictions. \vspace*{-9pt}

\subsection{Visualizing Combinatorial Interactions among Histone Modifications}
In order to interpret the combinatorial interactions among histone modifications, we present a visualization technique in Section 3.3. Figure~\ref{fig:dashboard} presents four visualization results from DeepChrome on four cell types with high AUC scores. Each visualization result is a heat-map which shows the histone modification combinatorial pattern that is best representative of high (label = $+1$) or low (label = $-1$) gene expression. Note that this is different than Section 5.2 where we validated the importance of bin positions in general, rather than the combinatorial interactions for a specific class. The values in the heatmaps are within the range $[0,1]$, representing how important a particular bin is for prediction of the class of interest. A threshold of $0.25$ was selected to filter ``active'', or important, bins that are most influential for a particular classification. We calculated the frequency count of active bins for each histone modification. Histone marks with high frequency counts ($>$ mean frequency count across all histone marks) are considered to be strongly affecting the gene expression to become high or low. As expected, we observe a relationship among promoter and structural histone modification marks (H3K4me3 and H3K36me3) for $47$ out of $56$ ($84\%$) cell-types when gene expression is high. Similarly, we observe an opposite trend with repressor marks (H3K9me3 and H3K27me3) showing combinatorial relationship for 50 out of 56 ($89\%$) cell-types, when gene expression is low. In other words, our model automatically learns that in order to classify a high or low gene expression, there needs to be high counts among promoter marks, or repressor marks, respectively. 

Next, we validated our visualization results with the findings in previous studies. Both of our baseline papers, \cite{cheng2011statistical} and \cite{dong2012modeling}, showed that there is a combinatorial correlation between H3K4me3 (promoter mark) and H3K36me3 (structural mark). This pattern can be seen in Figure~\ref{fig:dashboard} for high gene expression cases. Similarly, \cite{dong2012modeling} also reported a combinatorial correlation between promoter mark (H3K4me3) and distal promoter mark (H3K4me1), which is also validated by the DeepChrome visualization for $35$ out of $56$ cell-types ($62.5\%$). In addition, experimental studies have shown that these promoter marks play a role in the activation of genes, and this trend is seen in our visualization when the assigned label is $+1$.

Another combinatorial pattern that we noticed in the majority of cell-types ($89\%$, i.e $50$ out of $56$ cell-types) was that of H3K9me3 (heterochromatin repressor) and H3K27me3 (polycomb repressor) for low gene expression case (label=$-1$). We found this observation in multiple recent biological studies such as \cite{boros2014polycomb}. This study reported that these two repressor marks coexist and cooperate in gene silencing. With almost no expert knowledge, we were able to find and visualize this combination through DeepChrome. To our knowledge, none of the previous computational studies have reported this combinatorial effect between H3K9me3 and H3K27me3. In short, the DeepChrome visualization technique provides the potential to learn novel insights into combinatorial relationships among histone modifications for gene regulation.
\vspace*{-9pt}

\section{Discussion}
We have presented DeepChrome, a deep learning framework that not only accurately classifies gene expression levels using histone modifications as input, but also learns combinatorial relationships among these modification marks that regulate genes. We implement a Convolutional Neural Network based model, inspired from deep learning work in image recognition applications, and evaluate its performance on 56 cell types from the latest REMC dataset. To our knowledge, we are the first to implement deep learning on the task of gene expression classification using histone modification signals. 

DeepChrome outperforms state-of-the-art models using SVM and Random Forests for the target task over 56 cell-types (or tasks). In addition, we propose an optimization strategy to extract combinatorial relationships among histone modifications from the trained models. Our findings not only validate previous observations but also provide new insights for underlying gene regulation mechanisms that have been observed in recent experimental studies. We note that these insights are, for now, restricted to our literature search. Therefore, we provide the optimized histone pattern maps from the DeepChrome models for all 56 cell types for both cases of gene expression being classified as high and low online (\url{www.deepchrome.org}). We hope that biologists are able to utilize these results for drawing significant hypotheses on histone modification interactions that lead to gene activation or silencing. 

For future work, we would like to observe DeepChrome's performance on adding histone modifications to understand their combinatorial effects as well. We would also perform cross-cell predictions, where one model is trained on data from one cell type and predictions are made on the other cell types. Previous studies have reported that the correlations among histone modifications remain consistent across cell types. However, the decrease in performance (right tail in Figure~\ref{fig:perf}) for some cell types in our results suggests the potential to explore this further. Another plausible direction is to understand the effect of relationships among histone modifications for regulating individual genes. This can help biologists in designing `epigenetic drugs' that can manipulate histone modification marks and control the expression of a particular gene target.

In summary, DeepChrome opens multiple new avenues for studying and exploration of genetic regulation via epigenetic factors. This is made possible due to deep learning's ability to handle a large amount of existing data as well as to automatically extract important features and complex interactions, providing us with important insights. Techniques like DeepChrome hold the potential to bring us one step closer to properly investigating gene regulation mechanisms, which in turn can lead to understanding of genetic diseases.

\section*{Acknowledgements}
We would like to thank Dr. Mazhar Adli (Department of Biochemistry and Molecular Genetics, University of Virginia) for helpful discussions. We would also like to acknowledge the contribution of Marina Sanusi, our undergraduate research assistant, in developing the DeepChrome website.\vspace{-18pt}

\bibliography{DeepChrome-ECCB}
\bibliographystyle{natbib}
\end{document}